\documentclass{article} % For LaTeX2e
\usepackage{iclr2021_conference,times}

\usepackage{amsmath,amsfonts,bm}

\def\eqref#1{equation~\ref{#1}}
\def\1{\bm{1}}

\DeclareMathAlphabet{\mathsfit}{\encodingdefault}{\sfdefault}{m}{sl}
\SetMathAlphabet{\mathsfit}{bold}{\encodingdefault}{\sfdefault}{bx}{n}

\usepackage{hyperref}
\usepackage{url}
\usepackage{subcaption}
\usepackage{tabularx}
\usepackage{graphicx}
\usepackage{amsmath}
\usepackage{xcolor}
\usepackage{color, soul}
\title{Continuous Weight Balancing}

\iclrfinalcopy

\author{Daniel J. Wu \\
Department of Computer Science\\
Stanford University\\
Stanford, CA 94305, USA \\
\texttt{danjwu@stanford.edu} \\
\And
Avoy Datta \\
Department of Electrical Engineering\\
Stanford University\\
Stanford, CA 94305, USA \\
\texttt{avoy.datta@stanford.edu} \\
}

\begin{document}

\maketitle

\begin{abstract}
We propose a simple method by which to choose sample weights for problems with highly imbalanced or skewed traits. Rather than naively discretizing regression labels to find binned weights, we take a more principled approach -- we derive sample weights from the transfer function between an estimated source and specified target distributions. Our method outperforms both unweighted and discretely-weighted models on both regression and classification tasks. We also open-source our implementation of this method (https://github.com/Daniel-Wu/Continuous-Weight-Balancing) to the scientific community. 
\end{abstract}

\section{Motivation}
Real-world datasets are heterogenous and frequently skewed. Oftentimes, the data points of interest, such as disease-positive patients in medical datasets \citep{classimbalancemedicine}, are rare, and disproportionately outnumbered. Indeed, class imbalance is a well-known problem in machine learning \citep{classimbalance, classimbalance2}; imbalanced datasets will generally produce imbalanced models -- in the case of extremely imbalanced datasets, training will result in degenerate models which opt to ignore some rare classes \citep{classimbalancecnns}. These models can encode dataset-specific biases, and perform disproportionately better on highly represented data \citep{buolamwini2018gender, biassurvey}. There is a rich body of literature that aims to mitigate the negative effects of dataset imbalance \citep{classimbalancemethod1, classimbalancemethod2, classimbalancemethod3}; in practice, however, simple class balancing via reweighting is sufficient for many tasks. This method assigns sample weights inversely proportional to the amount of data in each class, thereby effectively upsampling poorly represented classes. By virtue of being simple to understand, easy to use, and effective, class reweighting is a well-worn wrench in the machine learning toolbox.

In this work, we consider a similarly simple and easy-to-use strategy to mitigate a broader type of dataset skew -- imbalance of a continuous trait. A common example of this is any regression task, in which the data may be concentrated in some section of the domain. These continuous traits need not just be labels; datasets may also be biased along the axis of some continuous feature or metadata -- i.e., patient age in medical datasets, or net worth in population studies. Correcting against these biases within datasets is a key step towards developing robust and unbiased models.

Our key contributions are threefold:
\begin{enumerate}
    \item We outline a method which approximates the underlying distribution of the continuous trait, and then chooses sample weights to adjust this distribution to an arbitrary target distribution. 
    \item We demonstrate the performance of our method on two canonical datasets -- the California housing prices dataset and the heart disease dataset -- and with three classes of models -- regression, random forest, and shallow neural networks.
    \item We provide an open-source and modular implementation of our method.
\end{enumerate}

\begin{figure}
    \centering
    \includegraphics[width=0.7\linewidth]{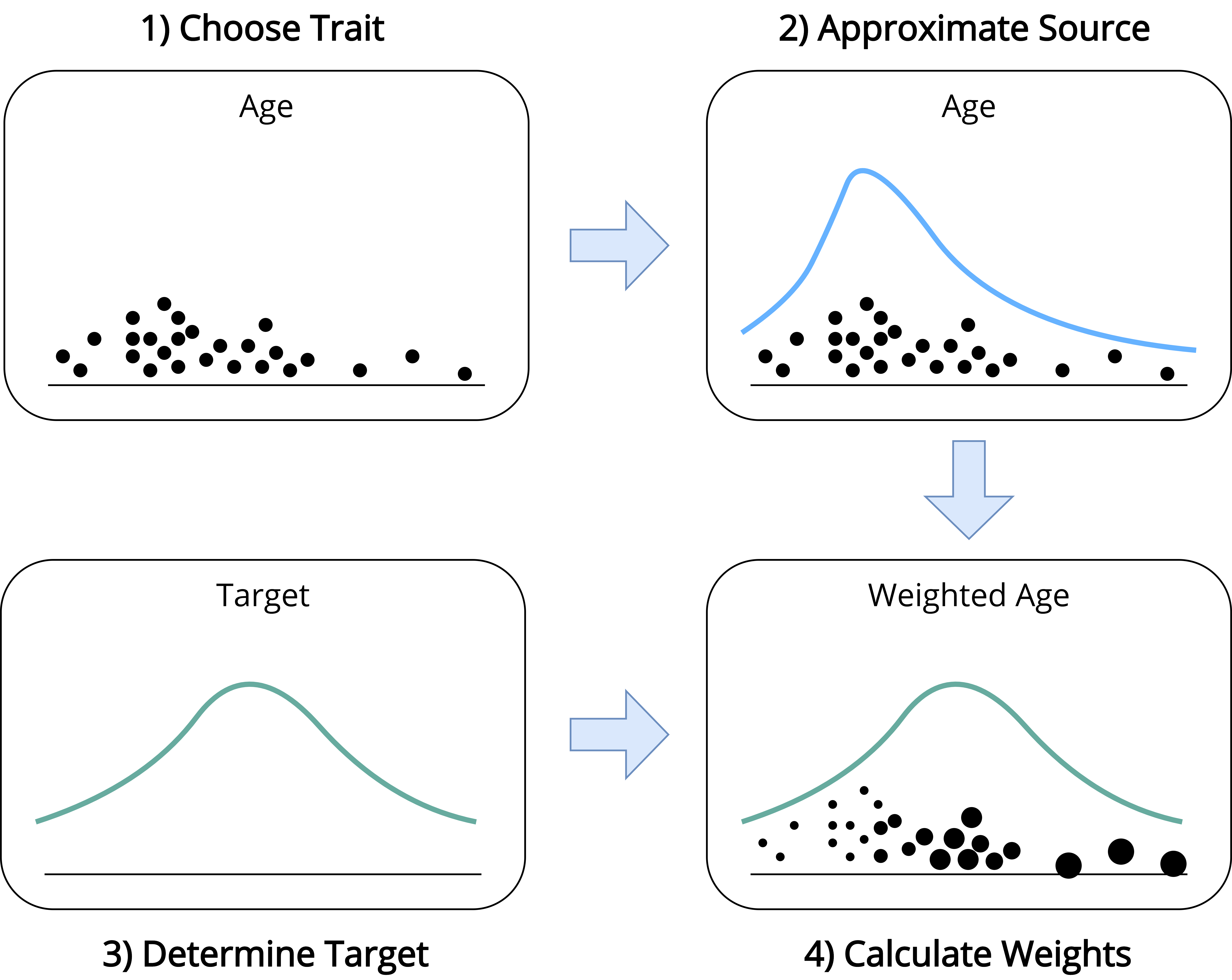}
    \caption{An illustration of continuous weighting in action.}
    \label{fig:demo}
\end{figure}

\section{Kernel Density Estimates}
Kernel Density Estimation \citep{davis2011remarks, parzen1962estimation} is a well-known method to evaluate the probability density of a random variable given some observed samples. Formally, let $x_1, x_2, ..., x_n$ be univariate samples drawn i.i.d. from a distribution with some density $f(x)$ at any given point $x$. We approximate this function $f$ with the kernel density estimator
\[
\hat{f}_h(x) = \frac{1}{n}\sum_{i=1}^nK_h(x-x_i) = \frac{1}{nh}\sum_{i=1}^nK\left(\frac{x-x_i}{h}\right)
\]
where $K$ is some non-negative kernel function, and $h>0$ is the bandwidth parameter which smooths the resultant estimate. By convenience, the standard normal density function $\phi$ is commonly used as the kernel function $K$.

The bandwidth parameter $h$ encodes a trade-off between the KDE's bias and variance; a common heuristic is Scott's rule \citep{scottsrule}, which, for a dataset of size $n$ with dimensionality $d$, sets the bandwidth $h$ as
$
h = n^{\frac{-1}{d+4}}
$.

\section{Method}
We outline a general and flexible framework to find weights which rebalance skewed datasets containing continuous features. Our method is a simple four-step procedure, which takes a dataset, and returns weights which map this data to some target distribution.

\paragraph{Choose a weight trait}
In our parlance, a weight trait is some continuous variable that captures an important feature of each data point; this is the variable we would like to weight based off of. This may be the label, a feature of the data point, or some metadata about each point. This trait may even be somewhat orthogonal to the modelling task. For example, the magnitude of each earthquake in earthquake datasets, or the volume of a given stock in financial datasets, may be traits which capture some notion of importance of a given data point, and for which we may want to weight. Choosing a weight trait may be a simple process of extracting some feature in the dataset or some corresponding data, or it may involve manual trait construction.

\paragraph{Approximate the source distribution} 
We approximate the empirical distribution of weight traits with a normal kernel density estimate, with bandwidth set by Scott's rule. This produces a smoothed estimate of the underlying data distribution, which is particularly useful in case of sparsely sampled or highly skewed traits. 

\paragraph{Determine a target distribution}
In this step, we determine the ideal distribution of the weight trait -- the distribution of the trait in our dataset that we would like to have, for one reason or another. Generally, this target distribution can be specified by some characteristics of the problem setting or dataset. For example, if the sample is skewed from the source population, it may be prudent to reweight traits (age, income, etc.) to match the source distribution. In a different vein, the problem may contain a trait that captures some notion of importance. For example, the market size of companies may be an important trait to weight with when assembling a portfolio that focuses on a certain company size.

\paragraph{Determine weights}
Once the source and target distributions are specified, the only remaining task is to find a set of weights which transforms the source distribution into the target distribution. One simple way to do this is to set the weight on each data point to the ratio between the source density and target density evaluated at that point. Formally, for a dataset $\{x_1, \dots, x_n\}$ with corresponding traits $\{t_1, \dots, t_n\}$, and an approximated source probability density $f_S$ and target probability density $f_T$, we calculate the corresponding weights as

\[
w_i = \frac{f_T(t_i)}{f_S(t_i)}
\]

\begin{figure}[ht]
\begin{subfigure}{.5\textwidth}
  \centering
  % include first image
  \includegraphics[width=0.8\linewidth]{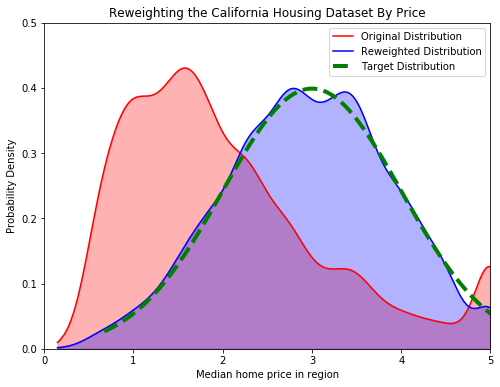}  
  \caption{California housing data reweighted to $\mathcal{N}(3, 1)$}
  \label{fig:data-ch}
\end{subfigure}
\begin{subfigure}{.5\textwidth}
  \centering
  % include second image
  \includegraphics[width=0.8\linewidth]{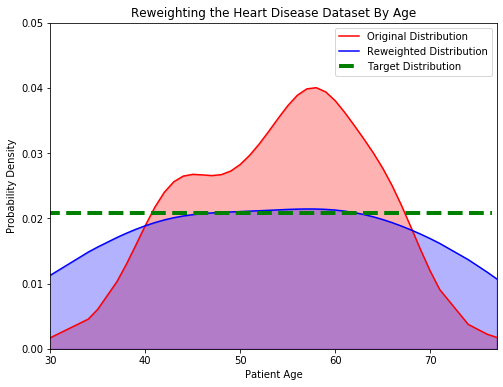}  
  \caption{The heart disease dataset reweighted to $\mathcal{U}(29, 77)$}
  \label{fig:data-hd}
\end{subfigure}
\caption{Our continuous reweighting method applied to our two datasets. Larger datasets (a) are easier to reweight than smaller datasets (b).}
\label{fig:datasets}
\end{figure}

\section{Experiments}
To demonstrate the practical efficacy of our method in correcting skewed datasets, we compare our weights against uniform and discretized weighting across several model classes.

\subsection{Datasets}
We run our experiments on two canonical datasets -- one regression, and one classification.
\paragraph{California Housing Dataset}
The California Housing dataset \citep{calihousing} contains the median housing prices of Californian census block groups in the 1990 census. It consists of 20,640 data points, where each data point contains 8 numeric attributes of houses in that block, and an accompanying median home price. These attributes are the population and median income of the block, and the latitude, longitude, average occupancy, average bedroom count, average room count, and median age of homes in the block. In common usage, the natural log of the home price is used as the label. For this dataset, we simply use the target variable -- the natural log of median housing prices -- as the weight trait. The dataset is skewed right, and we use a unit normal target distribution centered at 3 (Figure \ref{fig:data-ch}).

\paragraph{Heart Disease}

The Cleveland Heart Disease Database \citep{DETRANO1989304_heartdisease_uci}, containts the clinical information of 303 patients undergoing angiography. Each patient has 14 attributes, including demographic information such as age and sex, as well as condition-specific features such as rest ECG type and maximum heart rate. The target is a binary label indicating presence of heart disease. This dataset tests the limits of our method, as it is both extremely small and highly skewed. For this dataset we use age, which is slightly skewed left, as the weight trait, and we use a uniform target distribution across the domain as the weight trait (Figure \ref{fig:data-hd}). 

\begingroup
\begin{table}
\begin{tabularx}{\textwidth}{XXXXX}
\hline \hline

 \multicolumn{2}{c}{Model} & No weights & Discrete  &  CWB  \\ 
	\hline
\multicolumn{2}{l}{\textbf{California Housing (R2 score)}}  & & &\\
 \multicolumn{2}{l}{Random Forest regression}  & 0.5156 & 0.5042 & \textbf{0.5226}\\
 \multicolumn{2}{l}{Linear Regression}  & 0.0476 & 0.2892 & \textbf{0.3501}\\
 \multicolumn{2}{l}{Fully-connected network} & 0.4496 & 0.4237 & \textbf{0.5066}\\
 \hline 
 \multicolumn{2}{l}{\textbf{Heart Disease (AUROC)}}  & & & \\
 \multicolumn{2}{l}{Random Forest classification}  & 0.9554& 0.9593 & \textbf{0.9602}\\
 \multicolumn{2}{l}{Logistic Regression}  & 0.9261 & 0.9223 & \textbf{0.9242}\\
 \multicolumn{2}{l}{Fully-connected network} & \textbf{0.9375}& 0.9267 & 0.9324\\
\hline \hline

\end{tabularx}
\caption {\label{tab:tab_results} 
 Experiments on both datasets using continuous weight balancing (\textbf{CWB}), discrete balancing (\textbf{discrete}) and no weights. Metrics are reported on target out-of-sample subsets; patients under 60 for the heart disease dataset, and prices above 2 for the housing dataset.} 
\end{table}
\endgroup

\subsection{Evaluation}

We compare our continuous weights against \textit{no weighting} and \textit{discretized weighting}. For discretized weights, we first group training data into discrete bins based on a weight trait, then reweight the data in each bin until each bin has uniform representation in the training set. %For evaluation on the California housing dataset, we use an out-of-sample validation set 10\% the size of the dataset. Since Heart Disease is several orders of magnitude smaller in size, we used a larger out-of-sample set (20\%) to ensure adequate representation. 
We evaluate the R2 score for regression and AUROC for binary classification, on an out-of-sample subset of the data. Since our goal is to enable models to do well on underrepresented subsets of our data, we evaluate model performance on underrepresented subsets of the out-of-sample data. For the California housing dataset, we examine higher housing prices ($>2$), while for the heart disease dataset we examine lower age groups ($< 60$).

\subsection{Models}

We experiment with 3 classes of models: random forests, linear/logistic regression, and shallow neural networks. Our random forests use 100 estimators, and our neural networks are an ensemble of 10 feedforward neural networks, with two hidden layers containing 64 and 16 nodes, respectively. We use ReLU activations and apply dropout ($p=0.5$) between the two layers.

\section{Discussion}

Table \ref{tab:tab_results} outlines our method's performance. While performance gains are small in the smaller dataset (heart disease), our method significantly outperforms both discrete weights and no weighting across all models on the California Housing dataset.

In this work, we described a framework for continuous weight balancing, and assessed the performance of one simple way of doing so. While our method showed reasonable results, there are still many open questions about the best way to find continuous weights. What is the best way to approximate the source distribution? What classes of target distributions lead to the best performance, particularly with a given loss function? How do we strike the balance between skewed models, and models which memorize specific examples due to high magnitude weights? We hope that these and other questions may be answered in future work, in order to enable the development of robust models on skewed datasets.

\bibliography{iclr2021_conference}
\bibliographystyle{iclr2021_conference}

\end{document}